\def\year{2018}\relax
\documentclass[letterpaper]{article} 
\usepackage{aaai18}  
\usepackage{times}  
\usepackage{helvet}  
\usepackage{courier}  
\usepackage{url}  
\usepackage{graphicx}  
\usepackage{color}

\usepackage{tabularx,ragged2e,booktabs,caption}
\newcolumntype{C}[1]{>{\Centering}m{#1}}

\usepackage{amsmath}
\usepackage{algorithm}
\usepackage[noend]{algpseudocode}
\makeatletter
\def\BState{\State\hskip-\ALG@thistlm}
\makeatother
\usepackage{enumitem}

\begin{document}
%

\title{Checkpoint Ensembles: Ensemble Methods from a Single Training Process}
\author{Hugh Chen\\
University of Washington\\ 
Seattle, WA\\
\And Scott Lundberg\\ 
University of Washington\\ 
Seattle, WA
\And Su-In Lee\\
University of Washington\\ 
Seattle, WA
}
\maketitle
\begin{abstract}
We present the \emph{checkpoint ensembles} method that can learn ensemble models on a single training process. Although checkpoint ensembles can be applied to any parametric iterative learning technique, here we focus on neural networks.  Neural networks' composable and simple neurons make it possible to capture many individual and interaction effects among features.  However, small sample sizes and sampling noise may result in patterns in the training data that are not representative of the true relationship between the features and the outcome.  As a solution, regularization during training is often used (e.g. dropout). However, regularization is no panacea -- it does not perfectly address overfitting. Even with methods like dropout, two methodologies are commonly used in practice. First is to utilize a validation set independent to the training set as a way to decide when to stop training.  Second is to use ensemble methods to further reduce overfitting and take advantage of local optima (i.e. averaging over the predictions of several models).  In this paper, we explore checkpoint ensembles -- a simple technique that combines these two ideas in one training process.  Checkpoint ensembles improve performance by averaging the predictions from ``checkpoints'' of the best models within a \emph{single} training process.  We use three real-world data sets -- text, image, and electronic health record data -- using three prediction models: a vanilla neural network, a convolutional neural network, and a long short term memory network to show that checkpoint ensembles outperform existing methods: a method that selects a model by minimum validation score, and two methods that average models by weights. Our results also show that checkpoint ensembles capture a portion of the performance gains that traditional ensembles provide.
\end{abstract}

\section{Introduction}
Ensemble methods are learning algorithms that combine multiple individual methods to create a learning algorithm that is better than any of its individual parts \cite{Dietterich:2000:EMM:648054.743935}. The simplest methods are random initialization ensembles (RIE) which run the same model over the same data with different weight initializations. Ensemble methods have gained popularity because they can outperform any single learner on many datasets and machine learning tasks \cite{Krogh95neuralnetwork,Dietterich:2000:EMM:648054.743935,Ury}. 

However, when using deep learning models, ensemble methods are more challenging -- simply because training deep neural networks over large datasets takes a correspondingly large amount of computation time.  In the simplest case, a network is trained epoch by epoch using large datasets, iterating over the training data in order to calculate the gradient of the loss function, approaching an optimum by shifting parameters in the model (typically weights).  Eventually, after the score (e.g. loss, accuracy, etc.) on the validation set fails to improve after a number of epochs, the set of parameters that achieved the optimal performance on the validation set becomes the model to be evaluated on the test set.

\begin{figure}[ht]
\includegraphics[width=0.46\textwidth]{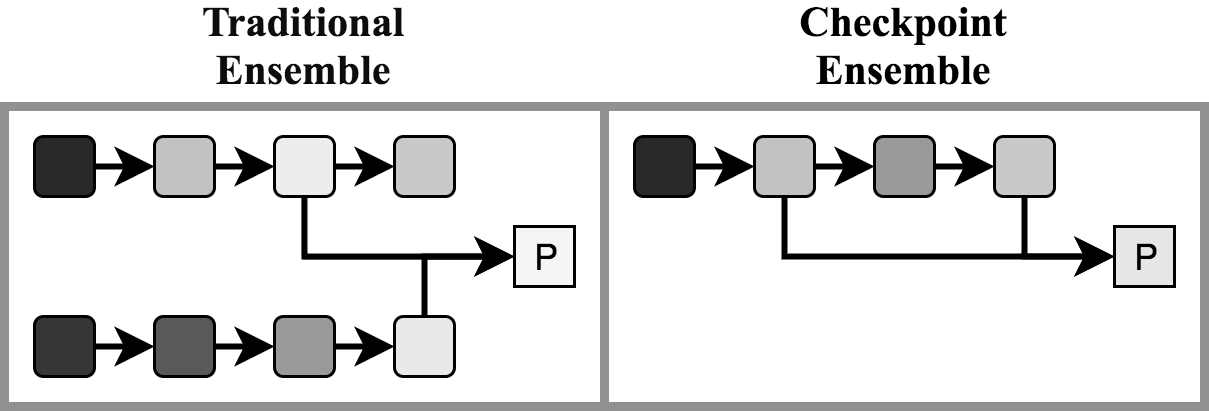}
\centering
\caption{The rounded boxes going from left to right represent models at each step of a particular training process (e.g. using gradient descent).  The shading represents validation score -- lighter shades represent a better score.  For either ensemble, we average the predictions from the best models to get the final prediction $P$.}
\label{checkpoint_fig}
\end{figure}

Checkpoint ensembles introduce the benefits of ensemble methods within a single network's training process by leveraging the validation scores.  Checkpointing refers to saving the best models in terms of a validation score metric across all epochs of a single training process.  Then, as in Figure \ref{checkpoint_fig}, based on the top scores we either combine the predictions of the top scoring models (checkpoint ensembles) or combine the models themselves by averaging (checkpoint smoothers).  Out of the single training process model averaging techniques considered in this paper -- minimum validation (MV), checkpoint smoothers (CS), and last $k$ smoothers (LKS) -- checkpoint ensembles show the best performance. It is worth noting that while checkpoint ensembles are particularly suitable to neural networks, they can easily be extended to any iterative learning algorithm.  Additionally, despite its performance and simplicity, checkpoint ensembles are surprisingly unexplored.  Ju et al. (2017) mentions checkpoint ensembles in an exploration of different methods for combining predictions with neural network ensembles, and Sennrich et al. (2017) used a checkpoint ensemble based off $N$ sequential epochs. \cite{Ju2017TheRP,uedin-nmt:2017}.

\section{Related Work}
For neural networks, one of the most commonly used single training process techniques for preventing overfitting is the \emph{minimum validation model selection (MV)} method which selects the model with the best validation score as the final model. As such, we used the MV method as a baseline in our experiments (see Experimental Results section).

The other two single training process methods are two different versions of what we call \emph{smoothers}: \emph{last k smoothers (LKS)} and \emph{checkpoint smoothers (CS)}.  Smoothing refers to averaging the weights of models, a natural parallel to averaging the predictions from models.  Utans (1996) explains that averaging parameters can be problematic because (1) different local minima may be found, and (2) a particular solution can be represented by different permutations of hidden nodes \cite{Utans96weightaveraging}.  Despite these problems, smoothers are still worthwhile comparisons because using a single training process may alleviate these problems.

LKS takes the best model in terms of validation score and averages the weights from the last $k$ prior epochs.  CS averages the weights from $k$ of the best models in terms of validation score. More formally LKS and CS can be implemented as:

\begin{enumerate}

\item Train neural networks in a normal fashion such that at epochs $1,2,\cdots,n$ we learn corresponding models $M=\{M_1, M_2,\cdots,M_n\}$ as well as validation scores $V=\{V_1,V_2,\cdots,V_n\}$, where each model $M_i$ has a set of weight parameters $W_i$.
\item Order $V$ to get $V_o=\{V_{(1)},V_{(2)},\cdots,V_{(n)}\}$ and the models $M_o=\{M_{(1)},M_{(2)},\cdots,M_{(n)}\}$ such that $M_j=M_{(i)}$ where $V_j=V_{(i)}$.  Depending on the validation score, the ordering $V_o$ may either be increasing or decreasing such that $V_{(1)}$ represents the optimal value.
\item Then impose an ordering on $M$ that we denote $M'=\{M'_{1},M'_{2},\cdots,M'_{k}\}$ with weights $W'_{1},W'_{2},\cdots,W'_{k}$, where $k$ is the number of models used for smoothing.  
\begin{enumerate}
\item For LKS, set $k=5$ and impose $M'_1=M_{(1)}=M_\ell$.  Then, $M'_2,\cdots,M'_k=M_{\ell-1},\cdots,M_{max(1,\ell-(k-1))}$.
\item For CS, we set $M'=\{M_{(1)},M_{(2)},\cdots,M_{(k)}\}$, with $k=min(a+5,b,n)$, where $a$ is the number of early stopping rounds, with $b$ s.t. $M_b=M_{(1)}$, and with $n$ the total number of epochs.  For the Experimental Results section, $a=10$.
\end{enumerate}
\item Return model $M_S$ with weights $W_S=\frac{1}{k}\sum_{i=1}^k W'_i$.
\end{enumerate}

In order to demonstrate that checkpoint ensembles capture a portion of the effect garnered from traditional ensemble methods we use \emph{random initialization ensembles (RIE)} for comparison as well.  For RIE, run models with different random initializations: $M^1=\{M_1^1,\cdots,M_{n_1}^1\},\cdots, M^k=\{M_1^k,\cdots,M_{n_k}^k\}$.  Denote a prediction on a sample point $x$ as $M(x)$. Then, predictions for the final model $M_{RIE}$ for a sample point $x_o$ is $M_{RIE}(x_o)=\frac{1}{k}\sum_{i=1}^k M_{(1)}^i(x_o)$, where $M_{(1)}^i$ is the best scoring model in terms of validation for training run $i$.  For the Experimental Results section,  $k=5$.

\subsection{Pseudocode}
For the following pseudocode we assume lower validation scores are better.  Pseudocode for predicting with the minimum validation (MV) method is as below:
\begin{algorithm}[H]
\caption{Predict with MV}\label{MV_alg}
\begin{algorithmic}[1]
\State models = nn.train(earlyStop,additionalParameters)
\Procedure{predMV}{models,x}
\State models.sort(by=``val scores'',order=``increase'')
\State \Return models[0].predict(x)
\EndProcedure
\end{algorithmic}
\end{algorithm}

Pseudocode for predicting with both last k smoother (LKS) and checkpoint smoothers (CS) is as below:
\begin{algorithm}[H]
\caption{Predict with LKS}\label{LKS_alg}
\begin{algorithmic}[1]
\State models = nn.train(earlyStop,additionalParameters)
\Procedure{predLKS}{models,x}
\State k = min(5, len(models))
\State models.sort(by=``epochs'',order=``decrease'')
\State models = models[len(models)-bestEpoch(models):]
\State model = nn.emptyModel(additionalParamters)
\State model.weights = avgWeights(models[:k])
\State \Return model.predict(x)
\EndProcedure
\end{algorithmic}
\end{algorithm}

\begin{algorithm}[H]
\caption{Predict with CS}\label{CS_alg}
\begin{algorithmic}[1]
\State models = nn.train(earlyStop,additionalParameters)
\Procedure{predCS}{models,x}
\State bestEpoch = bestEpoch(models)
\State k = min(earlyStop+5, bestEpoch, len(models))
\State models.sort(by=``val scores'',order=``increase'')
\State model = nn.emptyModel(additionalParameters)
\State model.weights = avgWeights(models[:k])
\State \Return model.predict(x)
\EndProcedure
\end{algorithmic}
\end{algorithm}

Pseudocode for random initialization ensembles (RIE) is as below:
\begin{algorithm}[H]
\caption{Predict with RIE}\label{rie_alg}
\begin{algorithmic}[1]
\State k = 5, i = 0, bestModelLst= []
\While{i $<$ k}
\State models = nn.train(earlyStop,additionalParameters)
\State models.sort(by=``val scores'',order=``increase'')
\State bestModelLst.append(models[0])
\State i = i+1
\EndWhile
\Procedure{predRIE}{bestModelLst,x}
\State \Return average(bestModelLst.predict(x))
\EndProcedure
\end{algorithmic}
\end{algorithm}

\section{Checkpoint Ensembles}
\subsection{Overview}
\begin{figure}[ht]
\includegraphics[width=0.46\textwidth]{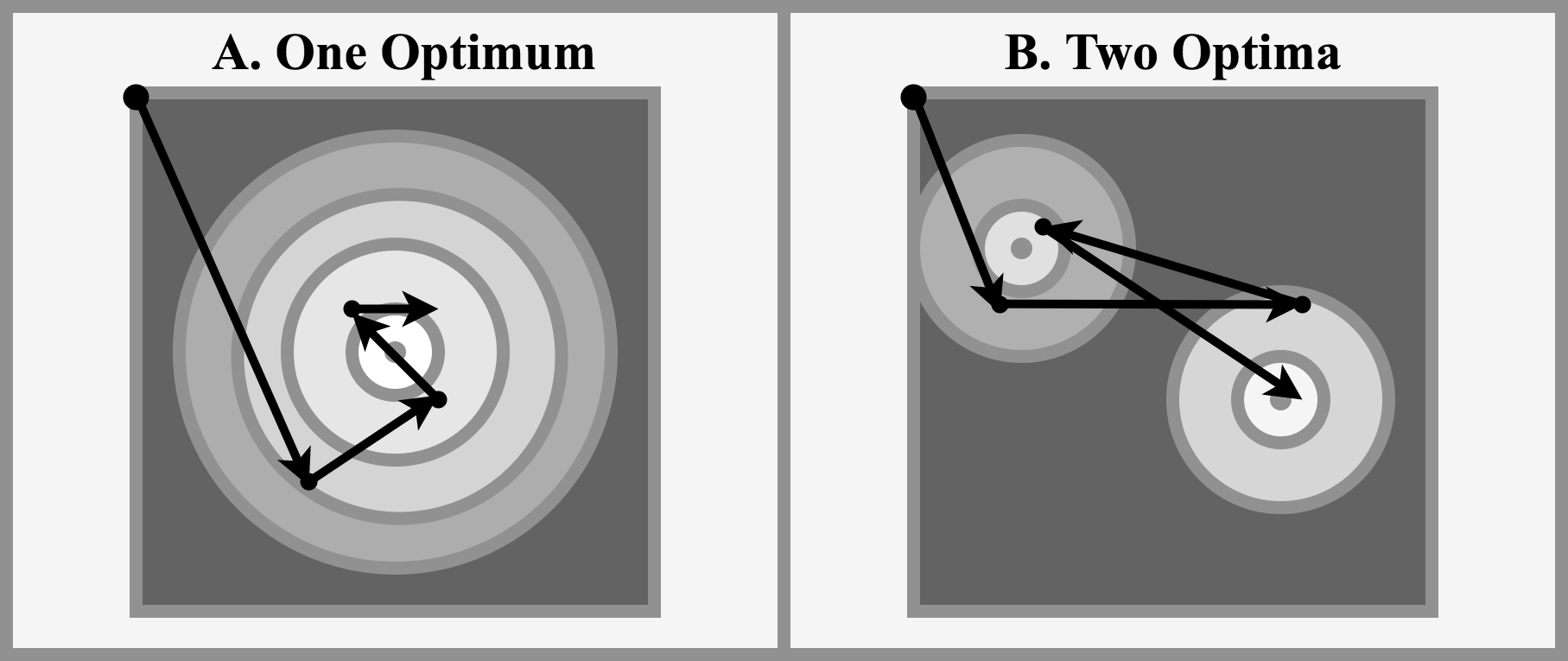}
\centering
\caption{Pictoral representation of scenarios for gradient descent: (A) when there is one optimal point and (B) when there are two local optima.  The shading represents the optimum in terms of our loss function, plotted against our (two) parameters.  The whiter the shade, the closer to optimal.  The arrows represent gradient descent.}
\label{optima_fig}
\end{figure}

We make two observations about the parameter space in relation to our loss function that serve as intuition.  First, on top of reducing overfitting, one intuition for why checkpoint ensembles (CE) would work well is that even at the end of training, the neural network may not end up exactly in a optimum point in the parameter space, as depicted in Figure \ref{optima_fig}A.  Even more crucially, as the network traverses the parameter space, the gradient and the learning rate may send the network around a single optimum (Figure \ref{optima_fig}A) or across multiple local optima (Figure \ref{optima_fig}B).  In either case, the models characterized by the weights at each of these epochs may be particularly confident when making predictions in unique regions of the space of all possible prediction problems (which we denote as $\Gamma$).  

To further elaborate, imagine model $M_1$ is good at prediction problems in the subspace $\alpha \in \Gamma$ and mediocre at the subspace $\beta \in \Gamma$ whereas $M_2$ is good at prediction problems in the subspace $\beta \in \Gamma$ and mediocre at the subspace $\alpha \in \Gamma$.  If this is the case, it stands to reason that both models will have high validation accuracy and therefore be \emph{checkpoints}.  Then, the final model $M_{CE}$, which is an ensemble of model $M_1$ with model $M_2$, will reflect the confidence $M_1$ has for the region $\alpha$.  Averaging $M_1$'s high probabilities for the predictions associated with region $\alpha$ with model $M_2$'s mediocre predictions for region $\beta$, result in $M_{CE}$ behaving like $M_1$ for region $\alpha$.  A parallel argument applies for $M_2$ and $\beta$.  Our resulting model $M_{CE}$ should then outperform either individual model $M_1$ or $M_2$.

The second observation is that scenarios like the ones in Figure \ref{optima_fig} are more common with moderately higher learning rates.  A model with a higher learning rate explores the parameter space more adventurously - likely visiting more valleys (optima) of the parameter space than a slightly lower learning rate.  These higher learning rates also come with another nice side effect; they generally require less epochs for the model to ``converge.''  This means that checkpoint ensembles potentially perform better with higher learning rates and thus need less epochs to reach their full potential than a neural network that chosen by minimum validation.

\subsection{Algorithm}
\begin{enumerate}
\item Train neural networks normally such that at epochs $1,2,\cdots,n$ we learn corresponding models $M=\{M_1, M_2,\cdots,M_n\}$ as well as validation scores $V=\{V_1,V_2,\cdots,V_n\}$.
\item Order $V$ to get $V_o=\{V_{(1)},V_{(2)},\cdots,V_{(n)}\}$  and the models $M_o=\{M_{(1)},M_{(2)},\cdots,M_{(n)}\}$ such that $M_j=M_{(k)}$ where $V_j=V_{(k)}$.  Depending on the validation score, the ordering $V_o$ may either be increasing or decreasing such that $V_{(1)}$ represents the optimal value.
\item Return model $M_{CE}$ where predicting on a sample point $x_o$ is $M_{CE}(x_o)=\frac{1}{k}\sum_{i=1}^k M_{(i)}(x_o)$. 
\end{enumerate}

To select $k$, a good heuristic is $k=min(a+5,b,n)$, where $a$ is the number of early stopping rounds, with $b$ such that $M_b=M_{(1)}$, and with $n$ the total number of epochs.  For the prediction problems in the Experimental Results section, we consider $a=10$.  We determined the heuristic $k$ by testing on the operating room data sets (see Experimental Results section).

\subsection{Pseudocode}
\begin{algorithm}[ht]
\caption{Predict with CE}\label{CE_alg}
\begin{algorithmic}[1]
\State models = nn.train(earlyStop,additionalParameters)
\Procedure{predCE}{models,x}
\State bestEpoch = bestEpoch(models)
\State k = min(earlyStop+5, bestEpoch, len(models))
\State models.sort(by=``val scores'',order=``increase'')
\State \Return average(models[:k].predict(x))
\EndProcedure
\end{algorithmic}
\end{algorithm}

\section{Experimental Results}
\subsection{Data} We consider the following three data sets:

\emph{Reuters} is a popular data set containing 11,228 newswires from Reuters with 46 topics - text categorization.

\emph{CIFAR-10} is a popular data set containing 60,000 32x32 color images with 10 classes - image classification.

\emph{Operating Room Data} contains 57,000 surgeries containing time series data and static summary information obtained under appropriate Institutional Review Board (IRB) approval.  After splitting surgeries into multiple time points, about 8,000,000 desaturation labels are present with about 120,000 positive examples.  In addition there are about 3,000,000 hypocapnia labels with about 240,000 positive examples.  Both of these dataset labels represent time series binary classification problems.

\subsection{Models}
We consider three neural network prediction models: vanilla neural networks, CNNs, and LSTMs.

Neural networks are well suited for checkpoint ensembles for two reasons: (1) training is a stochastic, iterative process, and (2) they are often applied on huge data sets where training time is expensive.  We implemented our networks in Python using Keras, a package that provides a convenient frontend to Tensorflow \cite{chollet2015keras,DBLP:journals/corr/LiptonKEW15}.

\emph{CNNs} utilize convolutions and have been applied with great success to image classification \cite{NIPS2012_4824}.

\emph{LSTMs} were introduced by Hochreiter and Schmidhuber as a variant on recurrent neural networks that avoid the vanishing gradient problem  \cite{Hochreiter:1997:LSM:1246443.1246450,hochreiter_1998}.

\subsection{Evaluation Metrics}
\emph{Accuracy} is a useful metric in its straightforward nature and ease of use.  In multilabel prediction problems for machine learning, a common approach is to predict the probabilities from the network with a given sample point and label the point based off the maximally probable class.  Then, accuracy is the percentage of labels that match the true labels for a given data set.

\emph{Area under the precision-recall (PR) curve} is considered as an evaluation metric. PR curves are widely used for binary classification tasks to summarize the predictive accuracy of a model. PR curves are popular for classification problems with imbalanced labels. True positives ($TP$) are positive sample points that are classified as positive whereas true negatives ($TN$) are negative sample points that are classified as negative.  Then, false positives ($FP$) are negative sample points that are classified as positive whereas false negatives ($FN$) are positive sample points that are classified as negative.  Precision is defined as $\frac{tp}{tp+fp}$ and recall is $\frac{tp}{tp+fn}$. The PR curve is plotted with precision (y-axis) for each value of recall (x-axis). In order to summarize this curve, it is conventional to use area under the curve (AUC) to measure prediction performance.

\subsection{Performance}
\subsubsection{Reuters}
The Reuters data set serves as a simple proof of concept.  We train a feedforward network structured as follows: 1,000 input nodes, 512 hidden node layer, ReLU activation, 0.5 dropout, 46 node layer, and Softmax.  We train until there was no improvement in validation score after ten epochs. We generate five models that vary only across each learning rate in the set $10^{-1.5},10^{-1.55},10^{-1.60},\cdots,10^{-3.5}$ (a total of 205 samples).  For each model we find the scores and epochs for MV, CE, CS, and LKS.  Then, for each learning rate we average across the MV predictions for the five models to get the RIE scores and add up the epochs to get the RIE epochs.

\begin{table}[ht]
\caption{Improvement over MV (Reuters)}
\label{table_reuters}
\begin{tabular}{ |p{1.1cm}||p{2.9cm}|p{2.9cm}| } 
 \hline
 Method & Difference CI 95\% & Difference p-value \\
 \hline
 \textbf{CE} & $\boldsymbol{[.0105,.0140]}$ & $\boldsymbol{2.2\times 10^{-16}}$\\ 
 \hline
 CS & $[.0017,.0048]$ & $8.3\times 10^{-5}$\\ 
 \hline
 LKS & $[-.0048,-.0016]$ & $8.4\times 10^{-5}$ \\
 \hline
 \textbf{RIE} & $\boldsymbol{[.0150,.0255]}$ & $\boldsymbol{1.6\times 10^{-9}}$ \\
 \hline
\end{tabular}
\\[10pt]
\caption*{We report p-values and confidence intervals for the one sample t-test to test the null hypothesis that we had zero difference from our baseline (MV).  The confidence intervals indicate the direction the performance moved relative to the baseline - positive values indicate better performance and negative values indicate worse performance.}
\end{table}

In Table \ref{table_reuters} we see that checkpoint ensembles (CE) and random initialization ensembles (RIE) consistently perform better than the baseline, minimum validation (MV), with RIE showing the best performance as one would expect.  The smoothers do not perform quite as well although there is a significant improvement in checkpoint smoothers (CS) over baseline that is not present for the last k smoother (LKS).  The improvement CS displays suggests that smoothing weights is a tenable approach under certain scenarios.

\begin{figure}[ht]
\includegraphics[width=0.45\textwidth]{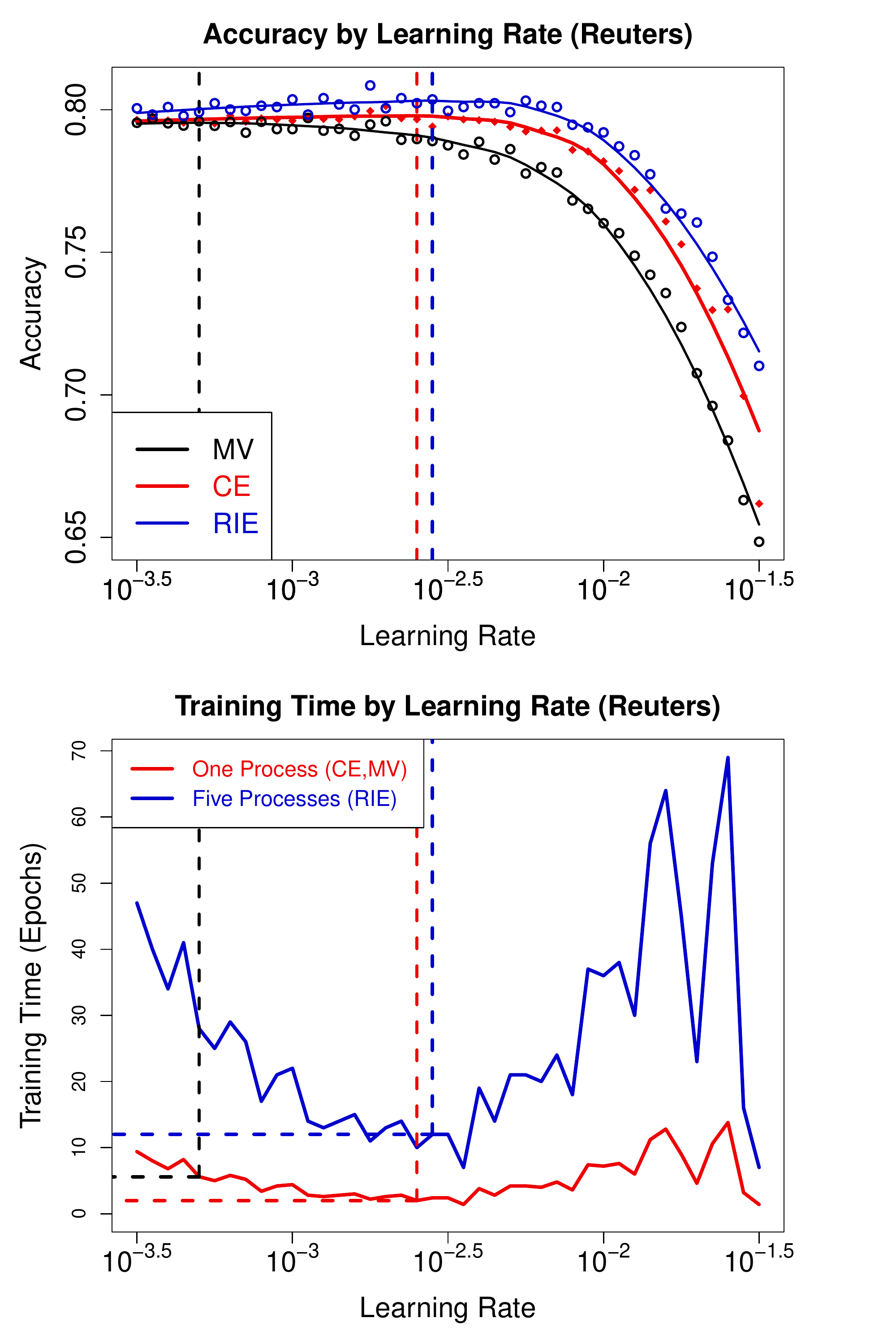}
\centering
\caption{Accuracy on the test set and epochs to convergence (i.e. number of sequential epochs to the maximum validation accuracy) for different learning rates.  We fit a spline to the accuracy and draw vertical lines through the maximum point on each of the splines.}
\label{reuters_mv_ce_rie}
\end{figure}

We averaged the accuracy and number of epochs for the five runs we used in Table \ref{table_reuters} for MV and CE for Figure \ref{reuters_mv_ce_rie} (we exclude LKS and CS for the sake of clarity, because they performed worse than CE).  For RIE, we already had a single estimate for the five runs across any particular learning rate.  First, we can notice that CE consistently outperforms selecting by the MV in terms of accuracy, and appears to capture some portion of the benefit RIE affords.  Secondly, we note that CE appears to converge at a higher learning rate than MV does.  

Now, looking at the epochs in Figure \ref{reuters_mv_ce_rie}, it is immediately obvious there is high variance in the epochs to the right of learning rate $10^{-2.5}$, because excessively high learning rates result in unreliable convergence.  At lower learning rates ($[10^{-3.5},10^{-2.5}]$), we see that the number of epochs to convergence grows as the learning rate decreases since the network moves through the space of parameters slowly.  Additionally, we observe that the maximum point for CE translates to approximately two epochs of training time whereas the maximum point for MV translates to five epochs of training time.  In this case, one should generally prefer CE because it requires less running time and achieves better performance than MV.  Comparing to RIE, we see that RIE takes about twelve epochs to converge under it's optimal learning rate compared to CE's two epochs.  This gain in convergence speed is less significant for the Reuters data set because it is small and straightforward; however for larger training data sets, a single epoch can easily take hours or days.  In these settings, CE might be a better choice of ensemble.  Finally, we observe that the optimal learning rate is very similar between CE and RIE.  Since the optimal learning rate between MV and RIE is quite different, another practical use for CE could be to tune the optimal learning rate for RIE.

\subsubsection{CIFAR-10}
The CIFAR-10 data set is more complicated than the Reuters data.  Since the data are images, we apply CNNs to solve the classification problem.  Our network is structured as follows: 32 neuron (3x3) Convolution layer, ReLU activation, 32 neuron (3x3) Convolution layer, ReLU activation, 2x2 MaxPooling, 0.25 Dropout, 64 neuron (3x3) Convolution layer, ReLU activation, 64 neuron (3x3) Convolution layer, ReLU activation, 2x2 MaxPooling, 0.25 Dropout, Flatten layer, 512 neuron Dense layer, ReLU activation, 0.5 Dropout, 10 neuron Dense layer, and Softmax.  We generate five models that vary across each learning rate in the set $10^{-2.5},10^{-2.55},10^{-2.60},\cdots,10^{-4.3}$ (a total of 185 samples).

\begin{table}[ht]
\caption{Improvement over MV (CIFAR-10)}
\label{table_cifar}
\begin{tabular}{ |p{1.1cm}||p{2.9cm}|p{2.9cm}| } 
 \hline
 Method & Difference CI 95\% & Difference p-value \\
 \hline
 \textbf{CE} & $\boldsymbol{[.0186,.0225]}$ &  $\boldsymbol{2.2\times 10^{-16}}$\\ 
 \hline
 CS & $[-.0180,-.0075]$ & $3.0\times 10^{-6}$ \\ 
 \hline
 LKS & $[-.0117,-.0079]$ & $2.2\times 10^{-16}$ \\
 \hline
 \textbf{RIE} & $\boldsymbol{[.0373,.0477]}$ & $\boldsymbol{2.2\times 10^{-16}}$ \\
 \hline
\end{tabular}
\\[10pt]

\caption*{We report p-values and confidence intervals for the one sample t-test to test the null hypothesis that we had zero difference from our baseline (MV).  The confidence intervals indicate the direction the performance moved relative to the baseline - positive values indicate better performance and negative values indicate worse performance.}
\end{table}

In Table \ref{table_cifar} we see significant improvements from CE and RIE as we did in the Reuters data set.  Additionally, we can see that the smoothers (LKS, and CS) are not good options for intra-process model averaging.  In fact, for the CIFAR-10 data set Checkpoint Smoothing (CS) has a generally negative effect, whereas it has a generally positive effect on the Reuters data set.

\begin{figure}[ht]
\includegraphics[width=0.45\textwidth]{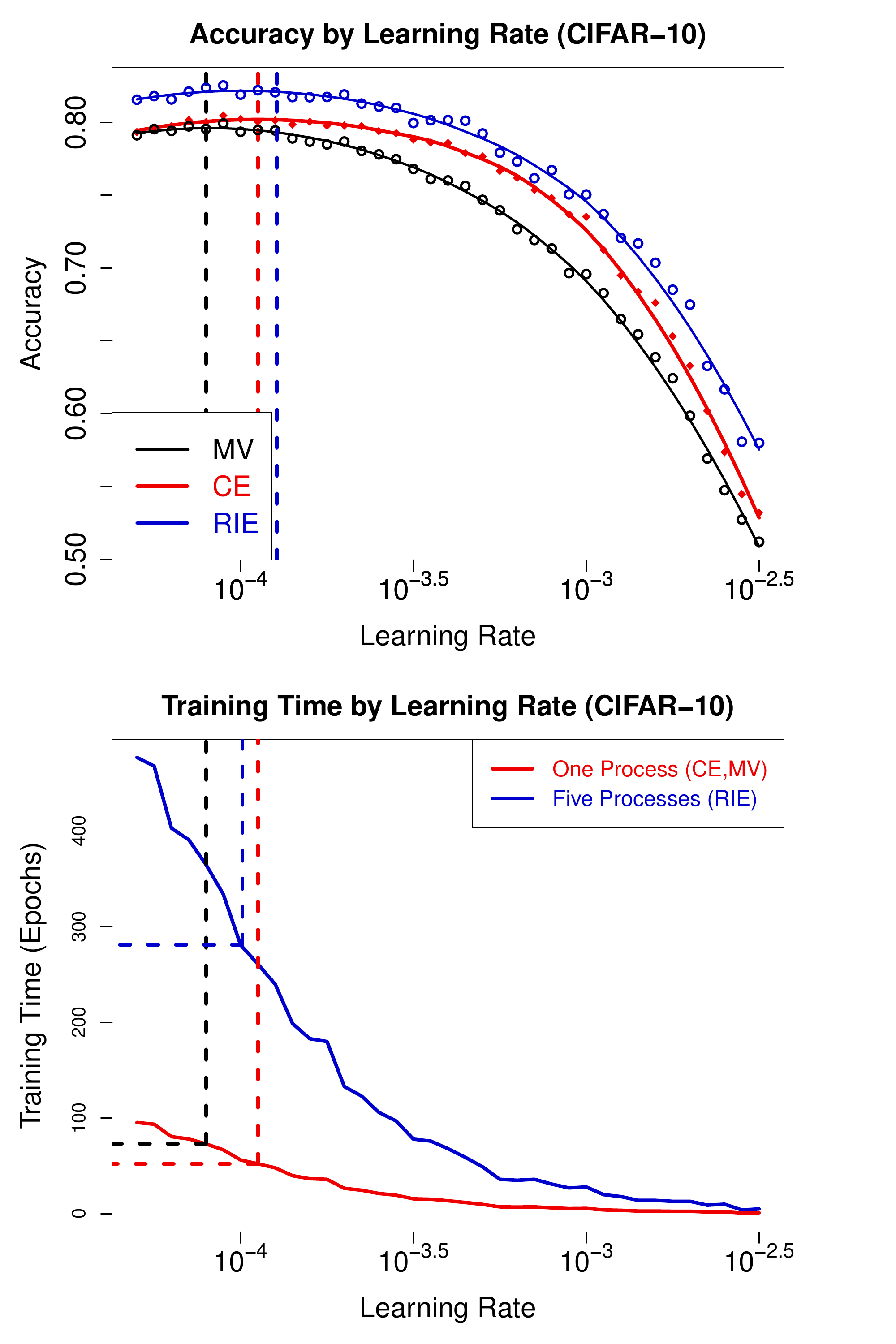}
\centering
\caption{Accuracy on the test set and epochs to convergence (i.e. number of sequential epochs to the maximum validation accuracy) for different learning rates.  We fit a spline to the accuracy and draw vertical lines through the maximum point on each of the splines.}
\label{cifar_mv_ce_rie}
\end{figure}

Once again, we average accuracy and number of epochs for the five runs we used in Table \ref{table_cifar} for MV and CE.  In Figure \ref{cifar_mv_ce_rie} there is a similar pattern emerging in terms of accuracy.  We see that CE outperforms MV in terms of accuracy and epochs, allowing for a bump in optimal performance as well as a reduction in training epochs required for the maximum accuracy - 70 to 50 (Figure \ref{cifar_mv_ce_rie}).  When inference is cheap CE should always be preferable to MV for training neural networks.  Then, comparing to RIE, we capture a portion of the benefit RIE provides with 50 epochs rather than 280 (Figure \ref{cifar_mv_ce_rie}).  In certain settings with extremely high training times, checkpoint ensembles could be preferable to RIE as well.  Finally, we observe that the optimal learning rate is very similar between CE and RIE once again, further supporting using CE as a cheap way to tune the optimal learning rate for RIE.

\subsubsection{Operating Room Data}
In this section we examine the performance of our methods on a significantly larger data set.  This data set contains both time series data and static summary information about patients in operating rooms. Since it contains time series data, we use an LSTM network for the two prediction tasks.  In order to keep the number of features manageable, we select the most common time series (e.g. SAO2, and ETCO2) features as well as natural static features (age, gender, height, weight, and ASACode).  Because this data set is so large, we only generate results for five learning rates.  Additionally, we use the AUC of the PR curve to measure our performance because PR curves are often used when the positive class is more interesting than the negative class, as is the case in these prediction problems. 

The first prediction task is oxygen desaturation, a medical condition that we define as the blood oxygen dropping below $92\%$.  The current state of the art method for this dataset is to XGBoost with pre-processed features (primarily exponential moving averages for the time series features).  Running XGBoost with early stopping, a step size of $0.02$ for $2000$ iterations on the subset of processed features yielded a model that achieved an AUC of the PR curve of $0.2306$ on the test set.

As a comparison, we ran an LSTM structured as follows: 41 input nodes, 400 LSTM nodes with $0.5$ recurrent dropout, 400 LSTM nodes with $0.5$ recurrent dropout, 0.5 dropout, and 1 output node with a sigmoid activation.  Rather than generating multiple models for each learning rate, we bootstrapped the test data fifty times in order to have a distribution of the possible prediction tasks.  Using these bootstrapped test sets, we calculated estimates for the test AUCs as well as standard deviations.

\begin{table}[ht]
\caption{Test AUC (OR Data - Desaturation)}
\label{table_desat}
\begin{tabular}{ |p{1.2cm}||p{.9cm}|p{.9cm}|p{.9cm}|p{.9cm}|p{.9cm}| } 
 \hline
 Learning Rate & 0.01 & 0.005 & 0.001 & 0.0005 & 0.0001 \\
 \hline
 \hline
 MV & 0.1057 & 0.2102 &0.2333 & 0.2261 & 0.2222 \\ 
 \hline
 MV ($\sigma$) & 0.0019 & 0.0033 & 0.0030 & 0.0029 & 0.0032 \\ 
 \hline
 CE & 0.1057 & 0.2136 & 0.2363 & 0.2323 & 0.2252 \\ 
 \hline
 CE ($\sigma$) & 0.0019 & 0.0034 & 0.0030 & 0.0030 & 0.0032 \\ 
 \hline
 Gain & 0.0000 & 0.0033 & 0.0030 & 0.0062 & 0.0030 \\ 
 \hline
 Gain ($\sigma$) & 0.0000 & 0.0006 & 0.0005 & 0.0004 & 0.0008 \\ 
 \hline
 Epoch  & 1 & 3 & 13 & 19 & 80 \\ 
 \hline
\end{tabular}
\\[10pt]
\caption*{This table reports AUC on the test set for desaturation.  Gain reports the improvement of CE over MV, and Epoch is the epoch where the model had the best validation score.}
\end{table}

In Table \ref{table_desat} LSTMs did indeed improve over XGBoost.  Furthermore, we see that our gain in performance from using checkpoint ensembles is significantly higher than the bootstrapped standard deviation.  Since this gain is past the AUC that the state of the art (XGBoost) achieves, it appears that checkpoint ensembles can offer significant performance gains in fairly difficult space of prediction problems.

Our next prediction task is hypocapnia, which is a medical condition defined as an end tidal CO2 of less than $35$ mmHg.  Using XGBoost on processed features with early stopping, a step size of $0.02$ for $2000$ iterations we found an AUC of $0.4369$ on the test set.

As a comparison, we ran an LSTM structured as follows: 41 input nodes, 200 LSTM nodes with $0.5$ recurrent dropout, 200 LSTM nodes with $0.5$ recurrent dropout, 0.5 dropout, and 1 output node with a sigmoid activation.  Once again, we bootstrapped the test data to obtain standard deviations for our estimates of performance.

\begin{table}[ht]
\caption{Test AUC (OR Data - Hypocapnia)}
\label{table_hypocapnia}
\begin{tabular}{ |p{1.2cm}||p{.9cm}|p{.9cm}|p{.9cm}|p{.9cm}|p{.9cm}| } 
 \hline
 Learning Rate & 0.01 & 0.005 & 0.001 & 0.0005 & 0.0001 \\
 \hline
 \hline
 MV & 0.1398 & 0.4059 &0.4279 & 0.4247 & 0.4256 \\ 
 \hline
 MV ($\sigma$) & 0.0011 & 0.0030 & 0.0030 & 0.0027 & 0.0029 \\ 
 \hline
 CE & 0.1398 & 0.4186 & 0.4365 & 0.4307 & 0.4283 \\ 
 \hline
 CE ($\sigma$) & 0.0011 & 0.0030 & 0.0031 & 0.0027 & 0.0027 \\ 
 \hline
 Gain & 0.0000 & 0.0127 & 0.0087 & 0.0060 & 0.0027 \\ 
 \hline
 Gain ($\sigma$) & 0.0000 & 0.0006 & 0.0005 & 0.0004 & 0.0007 \\ 
 \hline
 Epoch  & 1 & 7 & 11 & 10 & 63 \\ 
 \hline
\end{tabular}
\\[10pt]
\caption*{This table reports AUC on the test set for hypocapnia.  Gain reports the improvement of CE over MV, and Epoch is the epoch where the model had the best validation score.}
\end{table}

First of all, in Table \ref{table_hypocapnia} we see that our estimate for gain is once again significantly higher than its bootstrapped standard deviation.  Additionally we see that LSTMs do not generally improve over XGBoost in this setting, however utilizing checkpoint ensembles affords a performance gain that brings comparable performance at a learning rate of $0.001$.  Additionally, Table \ref{table_hypocapnia} shows that checkpoint ensembles generally perform better for learning rates on the higher side which is consistent with the Reuters and CIFAR-10 data sets.

\section{Discussion and Future Work}
We present and analyze a method to capture effects of traditional ensemble methods within a single training process.  Checkpoint ensembles (CE) provide the following benefits:
\begin{enumerate}
\item CE attains a significant amount of the benefit of traditional ensembles with significantly less training epochs.
\item CE attains optimal performance before minimum validation model selection, suggesting a necessity for less epochs.  Additionally, CE's optimal performance is higher than MV. 
\item CE can afford performance gains over minimum validation in simple neural networks, convolutional neural networks, and long short term memory networks.
\item CE offers a cheaper method to tune random initialization ensembles as we have seen that minimum validation is generally not a good approximation of the optimal parameter settings for RIE.
\end{enumerate}

The limitation of checkpoint ensembles is extra prediction time (capped out at a constant $a+5$ times the prediction time of a single model where $a$ is the number of early stopping rounds).  If prediction time is not a problem, CE should always be favored over MV.

As far as future work, checkpoint ensembles can be explored with more sophisticated schemes of ensembling such as the Bayes Optimal Classifier or the discrete Super Learner rather than a simple unweighted average \cite{vph07,Ju2017TheRP}.  Additionally, the utility of checkpoint ensembles could be explored in iterative learning techniques outside of neural networks.

\bibliography{ref}
\bibliographystyle{aaai}
\end{document}